\newif\ifniceformat
\niceformattrue     

\ifniceformat
\documentclass[11pt]{article}
\usepackage{amsmath, amssymb} 
\ifniceformat
\usepackage{amsthm}
\fi
\usepackage{bbm}                      
\usepackage{cancel}
\usepackage{mathtools}

\usepackage{graphicx}     
\usepackage{epstopdf}     
\usepackage{wrapfig}      
\usepackage{float}        
\usepackage{caption}      
\usepackage{subcaption}   
\usepackage{tikz}         
\usepackage{pgfplots}     
\usepackage{booktabs}

\usepackage{listings}                                
\usepackage[ruled,vlined,linesnumbered]{algorithm2e} 

\usepackage{xcolor}           
\usepackage{textcomp}         
\usepackage{csquotes}         
\usepackage[most]{tcolorbox}  

\usepackage{fullpage}      
\usepackage{etoolbox}      
\ifniceformat
\usepackage{enumitem}      
\fi
\usepackage{makecell}      
\usepackage{authblk}       
\usepackage{balance}       
\usepackage{multirow}

\usepackage{todonotes}     

\ifniceformat
\usepackage[round,sort,compress]{natbib} 
\else
\newcommand{\citep}[1]{\cite{#1}}
\newcommand{\citet}[1]{\cite{#1}}
\fi

\usepackage[
    colorlinks=true,
    citecolor=blue,
    linkcolor=blue,
    urlcolor=blue
]{hyperref}                                

\usepackage{auxFiles/symbolDef}
\usepackage{auxFiles/commands}

\makeatletter
    \@ifpackageloaded{xcolor}{}{\usepackage{xcolor}}
\makeatother
\makeatletter
    \@ifpackageloaded{needspace}{}{\usepackage{needspace}}
\makeatother

\newcommand{\uppercaseGreek}[1]{%
    \begingroup
    \ucmathlist\MakeUppercase{#1}%
    \endgroup
}

\newcommand{\ucmathlist}{%
    \def\alpha{\mathrm{A}}%
    \def\beta{\mathrm{B}}%
    \let\gamma=\Gamma
    \let\delta=\Delta
    \def\epsilon{\mathrm{E}}%
    \def\varepsilon{\mathrm{E}}%
    \def\zeta{\mathrm{Z}}%
    \def\eta{\mathrm{H}}%
    \let\theta=\Theta
    \let\vartheta=\Theta
    \def\iota{\mathrm{I}}%
    \def\kappa{\mathrm{K}}%
    \let\lambda=\Lambda
    \def\mu{\mathrm{M}}%
    \def\nu{\mathrm{N}}%
    \let\xi=\Xi
    \let\pi=\Pi
    \let\varpi=\Pi
    \def\rho{\mathrm{P}}%
    \def\varrho{\mathrm{P}}%
    \let\sigma=\Sigma
    \def\tau{\mathrm{T}}%
    \let\upsilon=\Upsilon
    \let\phi=\Phi
    \let\varphi=\Phi
    \def\chi{\mathrm{X}}%
    \let\psi=\Psi
    \let\omega=\Omega
}

\makeatletter
    \@ifpackageloaded{amsthm}{}{

        \usepackage{amsthm}
    }
\makeatother
\theoremstyle{plain}

\theoremstyle{definition}

\makeatletter
\def\renewtheorem#1{%
    \expandafter\let\csname#1\endcsname\relax
    \expandafter\let\csname c@#1\endcsname\relax
    \gdef\renewtheorem@envname{#1}
    \renewtheorem@secpar
}
\def\renewtheorem@secpar{\@ifnextchar[{\renewtheorem@numberedlike}{\renewtheorem@nonumberedlike}}
\def\renewtheorem@numberedlike[#1]#2{\newtheorem{\renewtheorem@envname}[#1]{#2}}
\def\renewtheorem@nonumberedlike#1{  
    \def\renewtheorem@caption{#1}
    \edef\renewtheorem@nowithin{\noexpand\newtheorem{\renewtheorem@envname}{\renewtheorem@caption}}
    \renewtheorem@thirdpar
}
\def\renewtheorem@thirdpar{\@ifnextchar[{\renewtheorem@within}{\renewtheorem@nowithin}}
\def\renewtheorem@within[#1]{\renewtheorem@nowithin[#1]}
\makeatother

\else
\documentclass[letterpaper, 10 pt, conference]{ieeeconf}
\IEEEoverridecommandlockouts 
\makeatletter
\let\NAT@parse\undefined
\makeatother
\fi

\ifniceformat

\usepackage{authblk}
\title{
Flatness-Preserving Residual Learning for Real-Time Tight Quadrotor Formation Flight
\thanks{
This work was supported in part by NSF Awards SLES-2331880, ECCS-2045834, ECCS-2231349, AFOSR Award FA9550-24-1-0102, DARPA TRS program under contract HR00112590145, ONR Award N000142512070, and ONR Award N000142512171.
}
}

\author[1]{Pei-An Hsieh\textsuperscript{\dag}}
\author[1]{Fengjun Yang\textsuperscript{\dag}}
\author[1]{Nikolai Matni}
\author[1]{M. Ani Hsieh}

\affil[$\dag$]{{\small Equal contribution}}
\affil[1]{{\small GRASP Laboratory, University of Pennsylvania, PA, USA}}

\else 

\title{\LARGE \bf Flatness-Preserving Residual Learning for Real-Time Tight Quadrotor Formation Flight}
\author{Pei-An Hsieh$^*$, Fengjun Yang$^*$, Nikolai Matni, M. Ani Hsieh%

\thanks{$^{*}$: Equal contribution. Authors are with the GRASP Laboratory, University of Pennsylvania, PA, USA. This work was supported in part by NSF Awards SLES-2331880, ECCS-2045834, ECCS-2231349, AFOSR Award FA9550-24-1-0102, DARPA TRS program under contract HR00112590145, ONR Award N000142512070, and ONR Award N000142512171.}%
}
\fi

\begin{document}
\maketitle

\begin{abstract}
Quadrotors flying in tight formations are severely affected by turbulent aerodynamic interactions, such as downwash, that can cause catastrophic collisions if left unmodeled. To compensate for these effects, we propose a physics-informed residual dynamics learning framework that captures complex aerodynamic interactions while ensuring the joint multi-quadrotor system remains differentially flat. We leverage this preserved flatness to design a computationally efficient feedback linearization controller that is easily tunable with linear control techniques and cancels aerodynamic disturbances via feedforward compensation. Hardware experiments demonstrate our framework reduces average tracking errors by $31\%$ compared to nominal baselines. Crucially, our lightweight approach matches the tracking performance of state-of-the-art nonlinear model predictive control (NMPC) while requiring an order of magnitude less computation. We are the first to show that stable, tight formation flight can be achieved with under 30 seconds of training data and a 5ms loop rate, unlocking high-fidelity aerodynamic compensation for compute-constrained flight stacks.

\ifniceformat
The video of our physical experiments can be found at \url{https://www.youtube.com/watch?v=uF26IkRFQMk}.
\fi
\end{abstract}



\section{Introduction}
Flying multiple quadrotors in close proximity poses significant challenges for controller design, as turbulent airflow generated by neighboring vehicles can degrade performance or even cause catastrophic collisions if left unaccounted for. Learning residual dynamics from flight data offers a powerful way to model these aerodynamic effects. To integrate general, unstructured residual models into the downstream control pipeline, state-of-the-art frameworks rely on NMPC \citep{chee2024flying}. However, NMPC demands substantial computational resources and scales poorly with the complexity of the learned model.

To overcome this computational bottleneck, structures, such as smoothness \citep{shi2019neural, shi2021neural} or specific dependencies \citep{spitzer2021feedback}, can be imposed on the learned residual to enable efficient feedback control. Building on this, we propose a physics-informed parameterization of the residual dynamics that strikes a balance between model fidelity and control efficiency by restricting our learned residual model to depend strictly on the joint positions and velocities of the quadrotor team. This ensures that the dynamics augmented by the learned residual are \textit{differentially flat}, a powerful property for quadrotor planning and control \citep{mellinger2011minimum}. Leveraging differential flatness, we design a computationally efficient controller that is easily tunable with standard linear control design techniques. Through simulation and hardware experiments, we show that our proposed controller achieves tracking performance on par with NMPC baselines \citep{chee2024flying} while requiring an order of magnitude less computation.

\begin{figure}
    \centering
    {\includegraphics[scale=0.25, trim = 0cm 0.4cm 0cm -0.3cm]{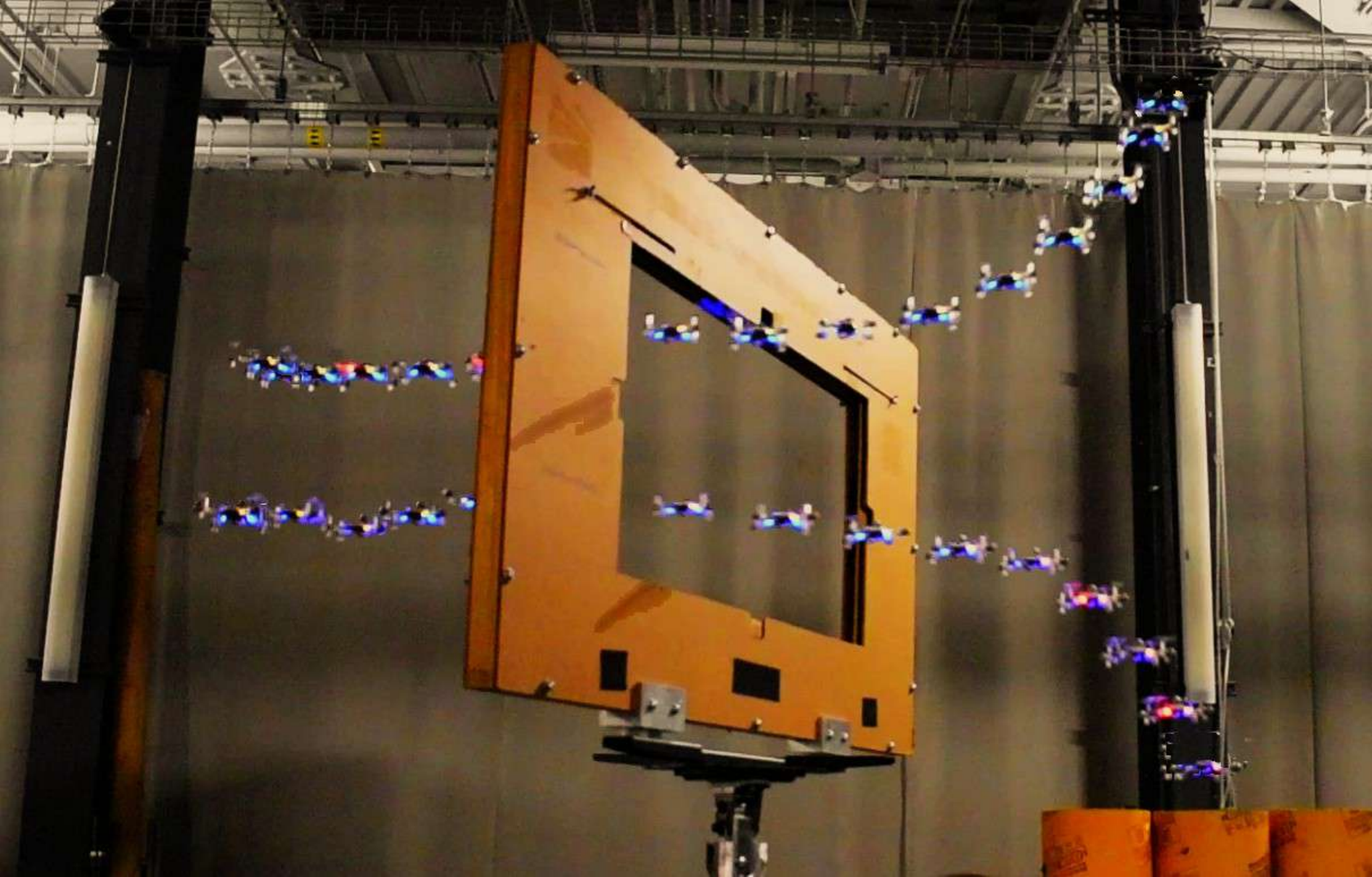}}
    \caption{ \small Time-lapse of two Crazyflie quadrotors using our proposed method merging from a  1.1 m \textit{stacked} formation into a tight 0.1 m separation to fly through a 0.4 m high window.
    }
    \label{fig:cover_image}
    \vspace{-3mm}
\end{figure}

\vspace{-1.5mm}
\subsection{Related Works}
\subsubsection{Residual Dynamics Learning for Quadrotors}
Prior works have successfully integrated learned residual dynamics to improve quadrotor tracking under aerodynamic disturbances. NeuralLander \citep{shi2019neural} and NeuralSwarm \citep{shi2021neural} enforce smoothness of the learned residual to establish stability guarantees for a cascaded controller. Unlike their approach, we utilize a more data-efficient, physics-informed parameterization and explicitly differentiates the learned residual to provide anticipatory feedforward attitude compensation to reduce phase lag during disturbances.
\ifniceformat
\citet{spitzer2021feedback} similarly restrict their residual dependency for feedback linearization, but only consider single-vehicle acceleration disturbances.
\else
Spitzer and Michael \cite{spitzer2021feedback} similarly restrict their residual dependency for feedback linearization, but only consider single-vehicle acceleration disturbances.
\fi
In contrast, we consider multi-quadrotor settings with a physics-informed residual parameterization and implement a different controller. Alternatively, the KNODE-MPC framework \citep{chee2024flying, chee2022knode, hsieh2025online} learns unconstrained, highly expressive residual dynamics that can be integrated into a downstream NMPC. While they achieve impressive performance, the optimization-based NMPC poses significant computational challenges for resource-constrained flight stacks. We directly compare our approach with this baseline in an identical hardware setup and show that our proposed controller matches their tracking performance while using $20\times$ less compute.

\subsubsection{Addressing Unmodeled Dynamics in Flatness-Based Control}
\ifniceformat
Flatness-preserving aerodynamic disturbances have been explored analytically in \citet{faessler2017differential} for quadrotors, who show that a specific, hand-derived aerodynamic drag model preserves differential flatness for a single quadrotor. Recently, \citet{yang2025learning} proposed a theoretical framework for learning flatness-preserving residual dynamics using structured parameterizations.
\else
Flatness-preserving aerodynamic disturbances have been explored analytically in Faessler et al. \cite{faessler2017differential} for quadrotors, who show that a specific, hand-derived aerodynamic drag model preserves differential flatness for a single quadrotor. Recently, Yang et al. \cite{yang2025learning} proposed a theoretical framework for learning flatness-preserving residual dynamics using structured parameterizations.
\fi
We leverage this framework and demonstrate its first successful hardware application for tight formation flight. Finally, rather than explicitly modeling residuals from data offline, an alternative is to actively estimate and reject the unmodeled dynamics online \citep{join2024flatness, hsieh2025online}. Integrating online estimation with our framework remains a promising direction for future work.

\vspace{-1mm}
\subsection{Contributions}
Our contributions are fourfold.
\begin{itemize}
    \item First, we show that the joint differential flatness and flat outputs of a multi-quadrotor team are preserved when the nominal rigid-body dynamics are augmented with a structured parameterization of the residual dynamics. We then explicitly derive the mappings to reconstruct states and inputs from flat outputs and their derivatives (Sec.~\ref{subsec:flatness-preservation-core}).
    \item Second, we leverage the flatness-preserving structure to design a residual dynamics learning framework that utilizes known physical structures to enable sample-efficient learning. (Sec.~\ref{sec:learning}).
    \item Third, we design a computationally efficient controller that leverages the learned residual dynamics to compensate for predicted aerodynamic disturbances (Sec.~\ref{subsec:FBL}). 
    \item Finally, we evaluate the proposed method in both simulation and hardware experiments and show that our proposed controller achieves similar tracking performance to an NMPC baseline while requiring an order of magnitude less computation (Sec.~\ref{sec:conclusion}).
\end{itemize}
To the best of our knowledge, this is the first time feedback linearization controllers with learned residuals have been applied to tight quadrotor formation flight.

\section{Problem Formulation}\label{sec:problem-formulation}
We consider a system of $N$ interacting quadrotors. Let $i \in \{1, \dots, N\}$ denote the index of an individual vehicle. The dynamics for agent $i$ are given by:
\begin{subequations}\label{eq:nominal-dynamics}
    \begin{align}
        \dot\vcp^i &= \vcv^i, \\
        m^i\dot\vcv^i &= u^i\mtR^i \vce_3 + m^i\vcg + \vcd^{a,i}, \label{eq:nom-trans-dyn}\\
        \dot\mtR^i &= \mtR^i \cpm{\vcomega^{i}}, \\
        J^i\dot\vcomega^i &= J^i\vcomega^i \times \vcomega^i + \vctau^{i} + \vcd^{\tau, i}. \label{eq:nom-rot-dyn}
    \end{align}
\end{subequations}
where $\vcp^i, \vcv^i \in \R^3$ are the position and velocity of vehicle $i$ in the world frame $\wdf$, and $\mtR^i \in \mathrm{SO}(3)$ its orientation. Its angular velocity $\vcomega^i$ is expressed in the body-fixed frame $\bdf$. The world frame follows an East-North-Up convention, with the three axes denoted as $\{\mathbf{e}_1, \mathbf{e}_2, \mathbf{e}_3\}$. The inputs are thrust $u^i \in \R$ and torque $\vctau^{i} \in \R^3$. The disturbances $\vcd^{a, i}$ and $\vcd^{\tau, i} \in \R^3$ arise from complex multi-agent aerodynamic effects. When the vehicles are sufficiently separated, these effects are small, and controllers can be designed for just the \textit{nominal} rigid-body dynamics that ignore such effects, i.e., treating $\vcd^{a, i} = \vcd^{\tau, i} = \mathbf{0}$.

However, when the quadrotors are flying in tight formations, complex aerodynamic effects like downwash inject significant disturbances (${\vcd^{a, i}, \vcd^{\tau, i} \gg \mathbf{0}}$) that can degrade tracking performance if left unaccounted for. Residual dynamics learning addresses this by using past trajectory data to learn approximate residual models $\hat \vcf^{a, i}$ and $\hat \vcf^{\tau, i}$ that predict these perturbations from a state-dependent feature vector $\vcxi$, yielding the \textit{augmented} acceleration dynamics for agent $i$\footnote{We let $\hat \vcf^{\tau, i}$ represent the residual torque in the world frame. It is thus mapped to the body frame via $(\mtR^i)^\top = \mtR^\bdf_\wdf$ in \eqref{eq:rot-dyn}.} as:
\begin{subequations}\label{eq:augmented-dynamics}
\begin{align}
    m^i\dot\vcv^i &= u^i\mtR^i \vce_3 + m^i\vcg + \hat \vcf^{a,i}(\vcxi),\label{eq:acc-dyn}\\
    J^i\dot\vcomega^i &= J^i\vcomega^i \times \vcomega^i + \vctau^{i} + (\mtR^i)^\top \hat \vcf^{\tau,i}(\vcxi).\label{eq:rot-dyn}
\end{align}
\end{subequations}

While the most general parameterization of $\hat \vcf^{a, i}$ and $\hat \vcf^{\tau, i}$ using the full multi-agent state maximizes expressivity, incorporating it into downstream planning and control is non-trivial and often requires computationally expensive NMPC \citep{chee2024flying}. To capture and reject unmodeled aerodynamic effects while minimizing the computational burden, we parameterize $\vcxi$ to depend only on the vehicles' positions and velocities. Crucially, this parameterization renders the joint augmented dynamics of the quadrotor team \textit{differentially flat}. This enables the use of flatness-based controllers that not only enjoy computational benefits but are also tunable with standard linear techniques (Sec.~\ref{sec:flatness-derivation}). Further, to mitigate potential generalization loss from this restricted expressivity, we parameterize the residual as the sum of a \rewrite{physics prior}{reduced-order model (ROM)} and a learned term (Sec.~\ref{sec:learning}). This formulation ultimately enables an efficient control pipeline, which we demonstrate experimentally in Sec.~\ref{sec:experiments}.

\section{Flatness-Preserving Residuals}\label{sec:flatness-derivation}
Informally, a system is differentially flat if its states and control inputs can be entirely reconstructed from a set of \textit{flat outputs} and their time derivatives \citep{fliess1995flatness}. This mapping between the flat outputs and system states establishes an equivalence between the nonlinear system and a linear one, thereby simplifying planning and control. In what follows, we first briefly review the differential flatness of a single quadrotor. We then introduce our parameterization of the residual dynamics and show that it renders the joint system differentially flat, allowing us to design controllers with simple linear design techniques. To reduce the notational burden, we drop the agent superscript in the derivations and present the formulation from the perspective of an ego-quadrotor, treating the states of all other agents as an exogenous parameter.

\subsection{Background on Quadrotor Flatness}
The nominal rigid-body quadrotor dynamics (i.e. Eq.~\eqref{eq:nominal-dynamics} with $\vcd^{a,i} = \vcd^{\tau, i} = \mathbf{0}$) are known to be differentially flat with respect to the flat outputs $\vcy = [\vcp, \psi]$, where $\psi$ is the yaw angle of the vehicle \citep{mellinger2011minimum}. We briefly recall how the state $[\vcp, \vcv, \mtR, \vcomega]$ and control inputs $[u, \vctau]$ can be reconstructed from $\vcy$ and its time derivatives.

Using the shorthand $\vcz:= \mtR \vce_3$ for the unit vector in the direction of the body $z$-axis and $\vca := \ddot \vcp$ as the acceleration, recall the translational dynamics \eqref{eq:nom-trans-dyn}
\begin{equation}\label{eq:nom-acc-expr}
    m\vca = u\vcz + m\vcg.
\end{equation}
Since $\vcz$ is a unit vector, the commanded thrust can be directly recovered from \eqref{eq:nom-acc-expr} as:
\begin{equation}\label{eq:nom-thrust-expr}
    u = m\norm{\vca - \vcg}.
\end{equation}
Rearranging this equation gives the body $z$-axis
\begin{equation}\label{eq:nom-body-z-expr}
    \vcz = \frac{m}{u}(\vca - \vcg).
\end{equation}
Crucially, because $\vca = \ddot{\vcp}$, both $u$ and $\vcz$ are entirely determined by the second derivatives of the flat outputs. The full orientation is then determined by the yaw angle $\psi$. Defining the heading vector $\bar{\vcx} = [\cos\psi, \sin\psi, 0]^\top$, the remaining body axes can be found as:
\begin{equation}\label{eq:nom-body-xy-expr}
    \vcy = \frac{\vcz \times \bar\vcx}{\norm{\vcz \times \bar\vcx}}, \quad \vcx = \vcy \times \vcz.
\end{equation}

To find the angular velocity, we must differentiate $\vcz$, which naturally introduces a dependence on the jerk ${\vcj:=\dddot \vcp}$. We define the world-frame angular velocity excluding the $z$-axis rotation as $\vcomega^\wdf_{xy}:= \vcz \times \dot{\vcz}$. Differentiating the acceleration equation \eqref{eq:nom-acc-expr} yields
\begin{equation}\label{eq:nom-jerk-expr}
    m\vcj = \dot u \vcz + u \dot\vcz.
\end{equation}
Taking the cross product of $\vcz$ with \eqref{eq:nom-jerk-expr} isolates the $x$ and $y$-axes angular velocities as
\begin{equation}
    \vcomega^\wdf_{xy} = \frac{m}{u} \vcz \times \vcj.
\end{equation}
Projecting this onto the body frame, we get that
\begin{equation}\label{eq:nom-rates-expr}
    \vcomega^\bdf_{xy} = \mtR^\top\vcomega^\wdf_{xy}= \frac{m}{u} \vce_3 \times \mtR^\top\vcj.
\end{equation}
The $z$-axis angular velocity $\omega_z$ is recovered from the yaw trajectory using standard quadrotor kinematics \citep{mellinger2011minimum}:
\begin{equation}\label{eq:nom-z-rate-expr}
    \omega_z = \frac{\omega_x (\vcz^\top \bar{\vcx}) + \left( \vcy^\top \dot{\bar\vcx} \right)}{\vcx^\top \bar{\vcx}},
\end{equation}
giving the full angular velocity $\vcomega = \vcomega^\bdf_{xy} + \omega_z \vce_3$. Finally, differentiating the angular velocity yields $\dot{\vcomega}$, which can be substituted into \eqref{eq:nom-rot-dyn} to recover the torque:
\begin{equation}\label{eq:nom-torque-expr}
    \vctau^\bdf = J\dot\vcomega + \vcomega \times J\vcomega.
\end{equation}

As a result, any sufficiently smooth trajectory of the flat outputs can be mapped to a dynamically feasible state and input trajectory via \eqref{eq:nom-acc-expr}-\eqref{eq:nom-torque-expr}. Thus, certain planning problems reduce to finding smooth flat output trajectories \citep{mellinger2011minimum}, and feedback controllers can be designed directly for the equivalent linear system (See e.g. \citet{fliess1995flatness}). However, introducing a residual can easily break this chain of algebraic substitutions. Next, we demonstrate that flatness is preserved provided the residual terms $\hat \vcf^a$ and $\hat \vcf^\tau$ depend strictly on the position and velocity of the quadrotor team.

\subsection{Flatness under Augmented Dynamics}\label{subsec:flatness-preservation-core}
We now consider the augmented dynamics \eqref{eq:augmented-dynamics}, which incorporate the learned residual terms. The acceleration and jerk equations become:
\begin{subequations}\label{eq:hod-expr}
\begin{align}
m\vca &= u\vcz + m\vcg + \hat \vcf^a(\vcxi),\label{eq:acc-expr}\\
m\vcj &= \dot u \vcz + u \dot\vcz + \dot {\hat \vcf}^a(\vcxi). \label{eq:jerk-expr}
\end{align}
\end{subequations}
As a result, if the residual feature vector $\vcxi$ included the thrust $u$, for example, solving for $u$ would become highly nontrivial. To preserve and reuse the chain of algebraic substitutions--and hence the flatness--of the nominal dynamics, we restrict our residual models $\hat \vcf^a(\vcxi)$ and $\hat \vcf^\tau(\vcxi)$ to depend exclusively on the joint kinematic state of the swarm:
\begin{equation}\label{eq:residual-form}
\vcxi = [\vcp^1, \vcv^1, \dots, \vcp^N, \vcv^N]^\top.
\end{equation}
We now show that in this case, we can reconstruct the state and control inputs from the same flat outputs $\vcy = [\vcp, \psi]$ via small adjustments to the nominal procedure.

First, we adapt \eqref{eq:nom-thrust-expr} and \eqref{eq:nom-body-z-expr} to recover the thrust and body $z$-axis by subtracting the acceleration residual:
\begin{subequations}\label{subeq:res-thrust}
\begin{align}
u &= \norm{ m(\vca - \vcg) - \hat \vcf^a(\vcxi) }, \label{eq:thrust-expr}\\
\vcz &= \frac{m(\vca - \vcg) - \hat \vcf^a(\vcxi)}{u}. \label{eq:body-z-expr}
\end{align}
\end{subequations}
The remaining body axes $\vcx$ and $\vcy$ are computed exactly as in \eqref{eq:nom-body-xy-expr} using the yaw angle. To find the angular velocity, we again isolate the $x$ and $y$ angular velocities from \eqref{eq:jerk-expr} and project them into the body frame, yielding
\begin{equation}\label{eq:rates-expr}
\vspace{-1mm}
\vcomega^\bdf_{xy} = \frac{1}{u} \vce_3 \times \mtR^\top \left( m\vcj - \dot{\hat{\vcf}}^a(\vcxi) \right),
\vspace{-0.5mm}
\end{equation}
where the residual derivative $\dot{\hat{\vcf}}^a$ can be computed via the chain rule, $\dot{\hat{\vcf}}^a(\vcxi) = \left[D_{\vcxi} \hat{\vcf}^a(\vcxi)\right] \dot{\vcxi}$. The Jacobian $D_{\vcxi} \hat{\vcf}^a$ can be obtained via automatic differentiation, and $\dot{\vcxi}$ consists of the swarm's concatenated velocities and accelerations. The $z$-axis rate $\omega_z$ is recovered identically to the nominal case as in \eqref{eq:nom-z-rate-expr}. Finally, incorporating the torque residual, the torque is found by rearranging \eqref{eq:rot-dyn}:
\begin{equation}\label{eq:torque-expr}
    \vctau = J\dot\vcomega + \vcomega \times J\vcomega - \mtR^\top \hat \vcf^\tau(\vcxi).
\end{equation}
Finally, because this algebraic mapping holds identically for every vehicle, the multi-quadrotor system is jointly flat with respect to the concatenated flat outputs of the individual vehicles. While we focus on two-quadrotor teams for this paper, this joint flatness provides an amenable structure for distributed flatness-based control in larger swarms \citep{yang2025scalable}.

\subsection{Flatness-based Controller Design}\label{subsec:FBL}
Assuming noisy measurements of $\vcp, \vcv$, and $\mtR$, we leverage flatness to design a cascaded feedback linearization controller that tracks a reference trajectory $\vcp_r$. We first compute a commanded acceleration via PID feedback on the position error:
\begin{equation}\label{eq:acceleration-command}
\vca = \vca_{r} - k_p(\vcp - \vcp_r) - k_i \int (\vcp - \vcp_r) dt - k_d(\vcv-\vcv_r),
\end{equation}
where $\vcv_r$ and $\vca_r$ are the reference velocity and acceleration. This commanded acceleration maps directly to the required thrust $u$ via \eqref{eq:thrust-expr}
\ifniceformat
\footnote{In our feedback implementation, we have measurements of the vehicle pose \(R\) and directly use the measured body \(z\)-axis, \(z=Re_3\), instead of computing \eqref{eq:body-z-expr}. Similarly, instead of \eqref{eq:thrust-expr}, we compute the desired thrust as $u = z^\top\left(m(a-g)-\hat f^a(\xi)\right).$ This expression is equivalent to \eqref{eq:thrust-expr} in the ideal case considered in our derivation, where the body \(z\)-axis is defined by \eqref{eq:body-z-expr}. In practice, the measured \(z\)-axis may not be perfectly aligned with the commanded force due to attitude tracking error, model mismatch, and measurement noise. We found the projection-based implementation to perform slightly better in hardware, potentially because it incorporates the measured thrust direction when the realized state deviates from the ideal flatness relation.}.
\else
.
\fi

To compute the desired angular velocity $\vcomega_r$, we differentiate \eqref{eq:acceleration-command} to find the corresponding jerk:
\begin{equation}\label{eq:jerk-command}
    \vcj = \vcj_{r} - k_p(\vcv - \vcv_r) - k_i(\vcp - \vcp_r) - k_d(\vca-\vca_r).
\end{equation}
The desired angular velocity $\vcomega_r$ is then recovered using \eqref{eq:rates-expr} and \eqref{eq:nom-z-rate-expr}. Because direct IMU acceleration measurements are often noisy and biased, we estimate the current vehicle acceleration $\vca$ from measured positions and velocities using an $\alpha-\beta-\gamma$ filter \citep{brookner1998tracking}. 

While the exact feedforward torque could be found via \eqref{eq:torque-expr}, evaluating the term $\dot\vcomega_r$ involves computing the second-order time derivative of the residual, $\ddot{\hat{\vcf}}^a(\vcxi)$. This exact evaluation adds computational overhead and is highly susceptible to noise amplification from high-order state derivatives. Therefore, we instead rely on a high-gain proportional controller on the angular velocity error to find the torque input
\begin{equation}\label{eq:controller-torque}
    \vctau = - K_\omega (\omega - \omega_r) - \mtR^\top \hat \vcf^\tau(\vcxi),
\end{equation}
where the torque residual $\hat{\vcf}^\tau(\vcxi)$ is incoporated as a feedforward compensation.

Finally, we note that although exact feedback linearization is possible via dynamic extension \citep{spitzer2021feedback}, it requires differentiating the acceleration residual $\hat \vcf^a$ to find $\ddot u$, which can filter out important low-frequency signals. We empirically found our approach to be significantly more robust for hardware deployment.

\section{Physics-Informed Residual Learning}\label{sec:learning}
To compensate for the potential generalization loss from preserving flatness, we inject strong inductive biases into our residual parameterization by explicitly encoding structural symmetries and introducing a \rewrite{physics-based downwash prior}{ROM of the downwash dynamics}.

Consider a pair of quadrotors with relative position $\Delta \vcp := \vcp^2 - \vcp^1 = [\Delta p_x, \Delta p_y, \Delta p_z]^\top$. We define the radial separation as $\Delta r:= \sqrt{\Delta p_x^2 + \Delta p_y^2}$ and the vertical separation as $\Delta z:= \Delta p_z$. Assuming near-hover flight for the upper quadrotor ($\vcz^1 \approx \vce_3$), we construct a local flow-field frame $\mathcal{F}$ centered at the upper quadrotor. Its orthonormal axes $\{\vce_x, \vce_y, \vce_z\}$ are defined as:
\begin{equation}
    \vce_z = -\vce_3, \;
    \vce_x = \vce_y \times \vce_z, \;
    \vce_y = \frac{\vce_z \times \Delta \vcp}{\norm{\vce_z \times \Delta \vcp}_2}.
\end{equation}
We denote the rotation matrix from $\mathcal{F}$ to the world frame $\wdf$ as $R^{\wdf}_{\mathcal{F}} \in SO(3)$. By projecting the relative kinematics into this frame, we form a rotationally equivariant feature vector $\vch_0 = [\Delta z, \Delta r]^\top$, while satisfying our flatness-preserving conditions.

Leveraging this invariant feature space, we parameterize the total residual dynamics $\hat \vcf(\vcxi)$ as the sum of a physics-based \rewrite{prior}{ROM} $\hat \vcf_d$ and a learned neural network $\hat \vcf_\Theta$:
\begin{equation}\label{eq:residual_model}
    \hat \vcf(\vcxi) = \hat \vcf_{d}(\vch_0) + \hat{\vcf}_{\Theta}(\vch_0),
\end{equation}
where each term can be broken into an acceleration component (e.g. $\hat \vcf_{d}^a$) and a torque component (e.g. $\hat \vcf_d^\tau$).

The \rewrite{physics prior}{ROM term} $\hat \vcf_{d}(\vch_0)$ is adapted from the pairwise downwash model in \citet{chee2024flying} and serves as a physically grounded foundation. It uses $\Delta z$ and $\Delta r$ to estimate the induced airflow velocity from the upper quadrotor \citep{bauersfeld2024robotics} and then applies standard drag equations to compute the resulting forces and torques on the lower vehicle \citep{jain2019modeling} (See details in Appendix~\ref{sec:physics-prior-details}).

To capture the remaining unmodeled dynamics, the learned component $\hat{\vcf}_{\Theta}(\vch_0) := [\hat{\vcf}^a_{\Theta^a}(\vch_0), \hat{\vcf}^\tau_{\Theta^\tau}(\vch_0)]^\top$ is parameterized as an $L$-layer feedforward neural network. For the acceleration residual:
\begin{equation}
\begin{aligned}
    \vch_{i+1} &= \sigma(W^a_i \vch_i + \vcb^a_i), \quad i=0, \dots, L-2\\
    \hat{\vcf}^a_{\Theta^a}(\vch_0) &= M_a [W^a_{L-1} \vch_{L-1} + \vcb^a_{L-1}]^\top,
\end{aligned}
\end{equation}
where $\Theta^a := \{W^a_l, \vcb^a_l\}_{l=0}^{L-1}$ are the learnable weights and biases, and $\sigma(\cdot)$ is a continuously differentiable activation function (e.g., $\tanh$) ensuring smoothness. The matrix $M_a = R^{\wdf}_{\mathcal{F}}/m$ transforms the network output from $\mathcal{F}$ to the world frame with the appropriate mass scaling. The torque residual $\hat \vcf^{\tau}_{\Theta^\tau}(\vch_0)$ is constructed identically with parameters $\Theta^\tau$ and scaling matrix $M_\tau = \inv{J} R^{\wdf}_{\mathcal{F}} / \|\inv{J}\|_F$, where $\|\cdot\|_F$ is the Frobenius norm.

Finally, to train the network parameters $\Theta = \{\Theta^a, \Theta^\tau\}$, we employ Knowledge-based Neural Ordinary Differential Equations (KNODE) \citep{jiahao2021knowledge, chee2022knode}, which avoids explicit finite differencing of noisy IMU measurements and the resulting biases. Given a dataset of $K$ trajectory samples $\mathcal{D} = \{(\vcx_k, \vcu_k)\}_{k=1}^K$, we optimize the weights using the Adam optimizer \citep{kingma2014adam} to minimize the weighted mean squared error between the integrated predictions $\hat{\vcx}_k(\Theta)$ and the ground-truth states:
\begin{equation}\label{eq:training_loss}
    \mathcal{L}(\Theta) := \frac{1}{K-1} \sum_{k=2}^K \|\hat{\vcx}_k(\Theta) - \vcx_k\|_{W_x}^2.
\end{equation}

\section{Experiment}\label{sec:experiments}
We validate our residual dynamics learning framework and flatness-based controller through simulation and hardware experiments. Our evaluations demonstrate three key results: (1) residuals learned with our flatness-preserving parameterization accurately capture complex downwash effects; (2) our proposed feedback linearization (FBL) controller (Section~\ref{subsec:FBL}) significantly reduces tracking errors by leveraging this learned residual; and (3) our approach achieves tracking accuracy comparable to a state-of-the-art nonlinear MPC \citep{chee2024flying} while reducing computation by an order of magnitude. Additionally, we show that our learned residual can provide feedforward compensation for standard SE(3) geometric controllers \citep{lee2010geometric}, highlighting its broad applicability.

To systematically validate these claims, we conduct simulation and hardware experiments that evaluate both the residual learning framework and the proposed FBL controller in a two-quadrotor \textit{stacked} formation flight, where the \textit{bottom} quadrotor flies directly within the turbulent wake of a \textit{top} quadrotor at a fixed vertical separation. Both vehicles execute straight, minimum-snap trajectories in the same direction, forcing the bottom quadrotor to constantly compensate for significant aerodynamic disturbances.

\ifniceformat
Further, to isolate the contributions from the \rewrite{physics-based prior}{reduced-order downwash model} and the learned neural network component \eqref{eq:residual_model}, we conduct ablation studies over four model fidelities:
\begin{itemize}
    \item \textbf{Nominal} ($\hat{\vcf}=0$): ignores aerodynamic effects.
    \item \textbf{ROM} ($\hat{\vcf}=\hat{\vcf}_d$): includes the ROM of the downwash dynamics but does not use learning.
    \item \textbf{Learned w/o ROM} ($\hat{\vcf}=\hat{\vcf}_\Theta$): learns the residual dynamics directly on top of the rigid-body quadrotor dynamics.
    \item \textbf{Learned + ROM} ($\hat{\vcf}=\hat{\vcf}_d + \hat{\vcf}_\Theta$): learns the residual on top of both the rigid-body quadrotor dynamics and the ROM of downwash dynamics.
\end{itemize}
Across the various experiments below, we demonstrate that both the \rewrite{prior}{ROM} and the learned neural network are crucial to the overall performance gains.
\else
Further, to isolate the contributions from the \rewrite{physics-based prior}{reduced-order downwash model} and the learned neural network component \eqref{eq:residual_model}, we conduct ablation studies over three model fidelities--\textbf{Nominal} (ignoring aerodynamic effects, $\hat{\vcf}=0$), \textbf{\rewrite{Physics Prior}{ROM}} ($\hat{\vcf}=\hat{\vcf}_d$), and \textbf{Learned \rewrite{}{}} (full model $\hat{\vcf}=\hat{\vcf}_d + \hat{\vcf}_\Theta$)--across the various experiments below, confirming that both components are crucial to the overall performance gains.\footnote{Due to space constraints, we defer the ablation that uses learning without ROM to our full technical report at \url{https://arxiv.org/pdf/2607.12275}
}
\fi

\subsection{Simulation Experiments}
\textit{Setup}: We simulate the dynamics \eqref{eq:nominal-dynamics} using the mass and moment of inertia of the Crazyflie quadrotors, subject to the downwash model from \citet{chee2024flying}. To emulate the discrepancy between our \rewrite{physics prior}{ROM} and true aerodynamic disturbances, we set the drag coefficient $C_D = 0.7$ for the simulated environment and $C_D = 0.3$ for the \rewrite{prior}{ROM}. Crucially, the simulated downwash model depends on variables outside our flatness-preserving parameterization \eqref{eq:residual-form} (e.g., vehicle orientations). Thus, this setup rigorously tests whether our strictly kinematic features can approximate complex, highly coupled aerodynamic interactions in the given scenario.

To capture residual forces and torques, we parameterize $\hat \vcf^a_{\Theta^a}$ and $\hat \vcf^\tau_{\Theta^\tau}$ as two separate neural networks with a single hidden layer of nine neurons each. For data collection, we utilize both the aforementioned \textit{stacked} formation, as well as a \textit{static top} formation, where the top quadrotor hovers while the bottom quadrotor executes a straight, constant-speed trajectory directly beneath it. The training dataset $\mathcal{D}$ comprises 18 seconds of flight data collected using the FBL controller augmented with the \rewrite{physics prior}{ROM} and consists of four scenarios: \textit{static top} and \textit{stacked} formations at vertical separations of 0.3 m and 0.4 m. The loss function \eqref{eq:training_loss} utilizes a diagonal weight matrix $W_x \in \mathbb{R}^{13 \times 13}$, with the final three elements set to $0.1$ and all others to $1$.

To test the controllers, we assume full, noise-free state observability. This isolates controller performance from the state estimator, a separation that is difficult to achieve in physical experiments. The FBL controller generates thrust and torque commands at 100 Hz, while the simulator integrates the dynamics via an RK45 method with a 10 ms time step. For the SE(3) controllers \citep{lee2010geometric}, the learned residual modifies the desired orientation via the exact same mapping defined in \eqref{eq:body-z-expr}.

\ifniceformat
\textit{Results}: We first evaluate the prediction error of our learned residual model both within and slightly outside its training distribution. We gather evaluation trajectories by flying two quadrotors in a stacked formation at vertical separations of $0.2, 0.3, 0.4$, and $0.5$m, sampling the bottom quadrotor's state at 100 Hz. Because we have access to the ground-truth downwash forces $\vcd^a$ and torques $\vcd^\tau$ in simulation, we define the acceleration and angular acceleration prediction errors for a given feature vector $\vcxi$ as:
\begin{equation*}
\begin{aligned}
    \vce^a(\vcxi) &= \frac{1}{m}\norm{\vcd^a(\vcxi) - \hat \vcf^a_\Theta(\vcxi)},\\ \vce^\tau(\vcxi) &= \norm{\inv{J}\left(\vcd^\tau(\vcxi) - \hat \vcf^\tau_\Theta(\vcxi)\right)}.
\end{aligned}
\end{equation*}
We report the root mean square error (RMSE) of these predictions in Table~\ref{tab:prediction_error}. Overall, incorporating the \rewrite{physics-based prior}{reduced-order downwash model} and the neural network residual yields significant performance gains. Our proposed learned model consistently achieves low acceleration prediction error, including on out-of-distribution (OOD) trajectories at $0.2$m and $0.5$m vertical separation. For angular acceleration, the learned model performs well in-distribution and at closer proximity ($0.2$m); however, at the larger $0.5$m separation, \rewrite{}{just using} the \rewrite{physics-prior model}{ROM} slightly outperforms it. This suggests that the neural network may mildly overfit to the highly turbulent close-proximity data, thereby marginally weakening its generalization to angular dynamics at larger, less perturbed distances. {On the other hand, learning the residual directly from data without leveraging the structured prior can achieve competitive translational acceleration error in some cases, but leads to substantially larger angular acceleration errors. Moreover, the \rewrite{physics-based prior}{ROM} provides the largest gains in regimes with stronger aerodynamic coupling, particularly at smaller vertical separations where extrapolation from limited data is more challenging. These results highlight the value of physics-based structure for improving generalization under limited data.}

\begin{table}[h]
\centering
\setlength{\tabcolsep}{4pt}
\begin{tabular}{@{}llcccc@{}}
\toprule
\textbf{Regime} & \textbf{$z$ [m]} & \textbf{Nominal} & \textbf{ROM} & \textbf{Learned w/o ROM} & \textbf{Learned + ROM} \\ \midrule
\multirow{2}{*}{In-Dist.}  
& 0.3 & 0.57 / 13.75 & 0.29 / 7.86 & 0.13 / 9.99 & \textbf{0.12} / \textbf{6.01} \\
& 0.4 & 0.52 / 11.55 & 0.27 / 6.80 & \textbf{0.09} / 10.09 & 0.10 / \textbf{6.02} \\ \midrule
\multirow{2}{*}{OOD} 
& 0.2 & 0.62 / 15.98 & 0.31 / 8.90 & 0.18 / 10.84 & \textbf{0.15} / \textbf{6.52} \\
& 0.5 & 0.47 / 9.84 & 0.25 / \textbf{6.05} & \textbf{0.08} / 10.78 & \textbf{0.08} / 6.36 \\ \bottomrule
\end{tabular}
\caption{Residual Prediction Error RMSE ($\vce^a$ / $\vce^\tau$).}
\label{tab:prediction_error}
\end{table}

To systematically evaluate the closed-loop tracking performance of the FBL and SE(3) controllers, we simulate an array of reference trajectories with varying vertical separations ($0.2$ to $0.6$m) and reference speeds ($0.2$ to $0.5$m/s). We measure performance using the 3D position RMSE and the maximum vertical tracking error ($z_{max}$), as downwash predominantly disrupts the vertical axis. Figure~\ref{fig:systematic-eval-fbl} visualizes the performance of the FBL controller across this sweep. Incorporating the physics-based flat downwash model immediately improves tracking errors compared to the Nominal baseline, particularly in the challenging close-proximity regime ($z=0.2$m). Adding the learned residual further reduces both the mean 3D position error and the maximum vertical tracking error, with the largest gains appearing in $z_{max}$. We report quantitative results in Table~\ref{tab:aggregated-control-fbl}, where the learned model consistently achieves the lowest or near-lowest tracking errors across both in-distribution and OOD separations. {The ablation without the \rewrite{physics prior}{ROM} shows that directly learning the residual can also improve closed-loop tracking, indicating that the FBL controller is relatively robust to approximation error in the learned model. However, incorporating the physics-structured prior in the learned model provides more reliable improvements at close separation, where aerodynamic coupling is strongest, and achieves the best overall 3D position tracking, which supports the prediction-error results in Table~\ref{tab:prediction_error}.}

In addition to the FBL controller, we use the learned residual to adjust the desired acceleration and computed torque in an SE(3) geometric controller~\citep{lee2010geometric}, with representative results reported in Table~\ref{tab:aggregated-control-se3}. The learned residual also improves SE(3) tracking performance over the Nominal and \rewrite{Physics Prior}{ROM} baselines, indicating that the learned correction is not tied to the FBL controller and can be incorporated into alternative control architectures. In particular, the gains persist across both the in-distribution and OOD separations evaluated here, supporting the broader applicability of the learned residual model.

\begin{figure}
    \centering
    \includegraphics[width=1.0\linewidth]{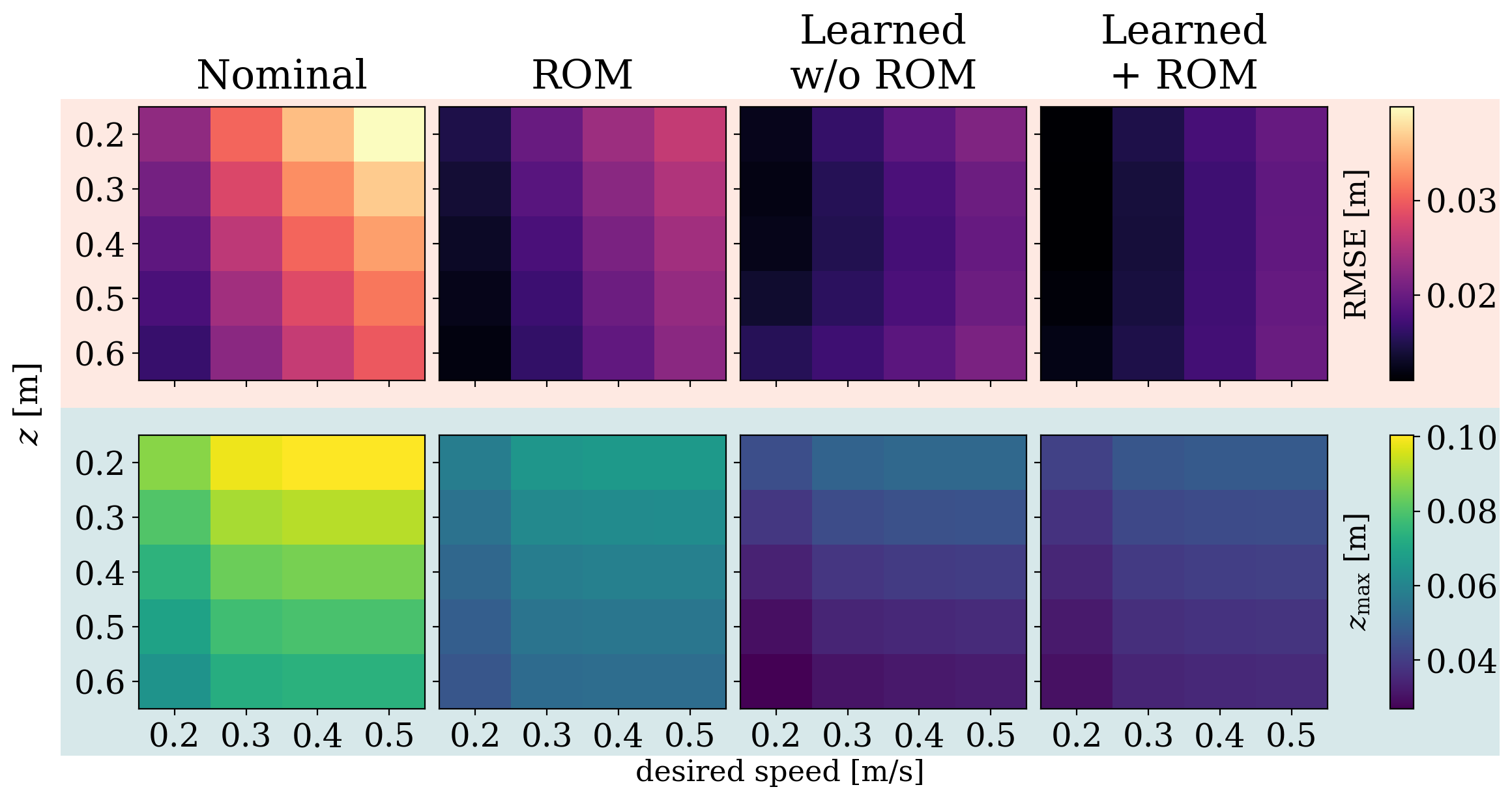}
    \caption{\small Closed-loop tracking performance (RMSE and $z_{max}$) of the FBL controller over varying reference speeds and vertical separations ($z$). The heatmaps illustrate the progressive reduction in tracking error when upgrading from the \rewrite{Nominal model (left) to the physics prior (center), and finally to the learned residual (right).}{Nominal model (left) to the ROM (center left), learned w/o ROM (center right) and finally to the residual learned with ROM (right).}}
    \label{fig:systematic-eval-fbl}
\end{figure}

\begin{table}[h]
\centering
\setlength{\tabcolsep}{4pt}
\begin{tabular}{@{}llcccc@{}}
\toprule
\textbf{Regime} & \textbf{$z$ [m]} & \textbf{Nominal} & \textbf{ROM} & \textbf{Learned w/o ROM} & \textbf{Learned + ROM} \\ \midrule
\multirow{2}{*}{In-Dist.}  
& 0.3 & 0.030 / 0.090 & 0.020 / 0.061 & \textbf{0.016} / 0.044 & \textbf{0.016} / \textbf{0.042} \\
& 0.4 & 0.028 / 0.083 & 0.019 / 0.057 & \textbf{0.016} / \textbf{0.038} & \textbf{0.016} / 0.039 \\ \midrule
\multirow{2}{*}{OOD} 
& 0.2 & 0.032 / 0.097 & 0.021 / 0.065 & 0.018 / 0.050 & \textbf{0.016} / \textbf{0.046} \\
& 0.5 & 0.026 / 0.077 & 0.018 / 0.054 & 0.017 / \textbf{0.034} & \textbf{0.016} / 0.036 \\ \bottomrule
\end{tabular}
\caption{Sim. Tracking Errors for the FBL Controller (RMSE / $z_{max}$) [m].}
\label{tab:aggregated-control-fbl}
\end{table}

\begin{table}[h]
\centering
\setlength{\tabcolsep}{4pt}
\begin{tabular}{@{}llcccc@{}}
\toprule
\textbf{Regime} & \textbf{$z$ [m]} & \textbf{Nominal} & \textbf{ROM} & \textbf{Learned w/o ROM} & \textbf{Learned + ROM} \\ \midrule
\multirow{2}{*}{In-Dist.} 
& 0.3 & 0.088 / 0.157 & 0.054 / 0.096 & 0.034 / 0.058 & \textbf{0.031} / \textbf{0.055} \\
& 0.4 & 0.079 / 0.140 & 0.048 / 0.084 & \textbf{0.026} / \textbf{0.044} & \textbf{0.026} / \textbf{0.044} \\ \midrule
\multirow{2}{*}{OOD} 
& 0.2 & 0.094 / 0.176 & 0.061 / 0.108 & 0.042 / 0.076 & \textbf{0.038} / \textbf{0.067} \\
& 0.5 & 0.070 / 0.124 & 0.042 / 0.074 & \textbf{0.021} / \textbf{0.031} & 0.022 / 0.035 \\ \bottomrule
\end{tabular}
\caption{Sim. Tracking Errors for the SE(3) Controller (RMSE / $z_{max}$) [m].}
\label{tab:aggregated-control-se3}
\end{table}

Finally, we evaluate our proposed controller against NMPC whose dynamics are augmented with the same learned residual model. Both controllers were tested head-to-head using the previous experimental setups, with results for mean 3D position error and maximum $z$ error reported in Table \ref{tab:nmpc_comparison}. Our flatness-based controller achieves tracking performance on par with the NMPC. Moreover, while the NMPC was implemented in Acados \citep{verschueren2022acados}, compiled into C code, and utilized real-time iteration, our JIT-compiled PyTorch-based controller still achieved a $20\times$ reduction in solve time. This efficiency suggests broader applicability for computationally constrained hardware.

\begin{table}[h]
\centering
\setlength{\tabcolsep}{4pt} 
\begin{tabular}{@{}clcc@{}}
\toprule
\textbf{$z$ [m]} & \textbf{Ctrl.} & \textbf{Error (RMSE/}$z_{max}$\textbf{)} & \textbf{Avg. Solve Time[ms]}  \\ \midrule
\multirow{2}{*}{0.3} & Ours & \textbf{0.016} / 0.042 & \textbf{0.99}  \\
 & NMPC & 0.020 / \textbf{0.032} & 27.6  \\ \midrule
\multirow{2}{*}{0.5} & Ours & \textbf{0.016} / 0.036 & \textbf{1.3}  \\
 & NMPC & 0.018 / \textbf{0.019} & 26.2  \\ \bottomrule
\end{tabular}
\caption{Benchmark against NMPC \citep{chee2024flying}}
\label{tab:nmpc_comparison}
\end{table}
\vspace{-2mm}
\else
\textit{Results}: We first evaluate the prediction error of our learned residual model both within and slightly outside its training distribution. We gather evaluation trajectories by flying two quadrotors in a stacked formation at vertical separations ($z$) of $0.2, 0.3, 0.4$, and $0.5$m, sampling the bottom quadrotor's state at 100 Hz. Because we have access to the ground-truth downwash forces $\vcd^a$ and torques $\vcd^\tau$ in simulation, we define the acceleration and angular acceleration prediction errors for a given feature vector $\vcxi$ as:
\begin{equation*}
\begin{aligned}
    \vce^a(\vcxi) &= \frac{1}{m}\norm{\vcd^a(\vcxi) - \hat \vcf^a_\Theta(\vcxi)},\\ \vce^\tau(\vcxi) &= \norm{\inv{J}\left(\vcd^\tau(\vcxi) - \hat \vcf^\tau_\Theta(\vcxi)\right)}.
\end{aligned}
\end{equation*}
We report the Root Mean Square Error (RMSE) of these predictions in Table~\ref{tab:prediction_error}. Overall, incorporating the \rewrite{physics-based prior}{ROM} and the neural network residual yields significant performance gains. Our proposed learned model consistently achieves the lowest acceleration prediction error, even on out-of-distribution (OOD) trajectories ($0.2$m and $0.5$m). For angular acceleration, the learned model performs well in-distribution and at closer proximity ($0.2$m); however, at a larger $0.5$m separation, the \rewrite{physics prior model}{ROM} slightly outperforms it. This suggests that the neural network may slightly overfit to the highly turbulent, close-proximity data, thereby marginally weakening its generalization to angular dynamics at larger, less perturbed distances.

\begin{table}[h]
\centering
\setlength{\tabcolsep}{4pt}
\begin{tabular}{@{}llccc@{}}
\toprule
\textbf{Regime} & \textbf{$z$ [m]} & \textbf{Nominal} & \textbf{ROM} & \textbf{Learned} \\ \midrule
\multirow{2}{*}{In-Dist.}  & 0.3 & 0.57 / 13.75 & 0.29 / 7.85 & \textbf{0.12} / \textbf{6.05} \\
& 0.4 & 0.52 / 11.55 & 0.27 / 6.79 & \textbf{0.10} / \textbf{6.04} \\ \midrule
\multirow{2}{*}{OOD} & 0.2 & 0.62 / 15.98 & 0.31 / 8.89 & \textbf{0.15} / \textbf{6.57} \\
& 0.5 & 0.47 / 9.84 & 0.25 / \textbf{6.05} & \textbf{0.08} / 6.37 \\ \bottomrule\end{tabular}
\caption{Residual Prediction Error RMSE ($\vce^a$ / $\vce^\tau$).}
\label{tab:prediction_error}
\end{table}

To systematically evaluate the closed-loop tracking performance of the FBL controllers and SE(3) controllers, we simulate an array of reference trajectories with varying vertical separations (0.2 to 0.6 m) and reference speeds (0.2 to 0.5 m/s). We measure performance using the 3D position RMSE and the maximum vertical tracking error ($z_{max}$), as downwash predominantly disrupts the vertical axis. Figure~\ref{fig:systematic-eval-fbl} visualizes the performance of the FBL controller across this sweep. We note that incorporating the \rewrite{physics-based flat downwash model}{ROM} immediately improves tracking errors compared to the Nominal baseline, particularly in the challenging close-proximity regime ($z=0.2$ m). Incorporating the learned residual model yields additional improvement in both mean 3D position error and maximum vertical error. We show quantitative results in Table~\ref{tab:aggregated_control} for two representative vertical separations (errors averaged over all reference speeds).  We note that while the improvement in RMSE is mild at larger separations ($z=0.5$ m), introducing the learned residual decreases the maximum vertical tracking error by  $32\%$ compared to just using the \rewrite{physics-based model}{ROM}. Further, while earlier metrics showed slight degradation in torque predictions of the learned module when the states are out-of-distribution, the acceleration accuracy proves more than sufficient to drive robust closed-loop improvements. In addition to the FBL controller, we use the learned residual to adjust the desired acceleration and computed torque in an SE(3) geometric controller \citep{lee2010geometric}. We observe that the SE(3) controller can also leverage the learned residual models to improve tracking performance, showing that our learned residual is not controller-dependent and has wider applicability.

\begin{figure}
    \centering
    \includegraphics[width=1.0\linewidth]{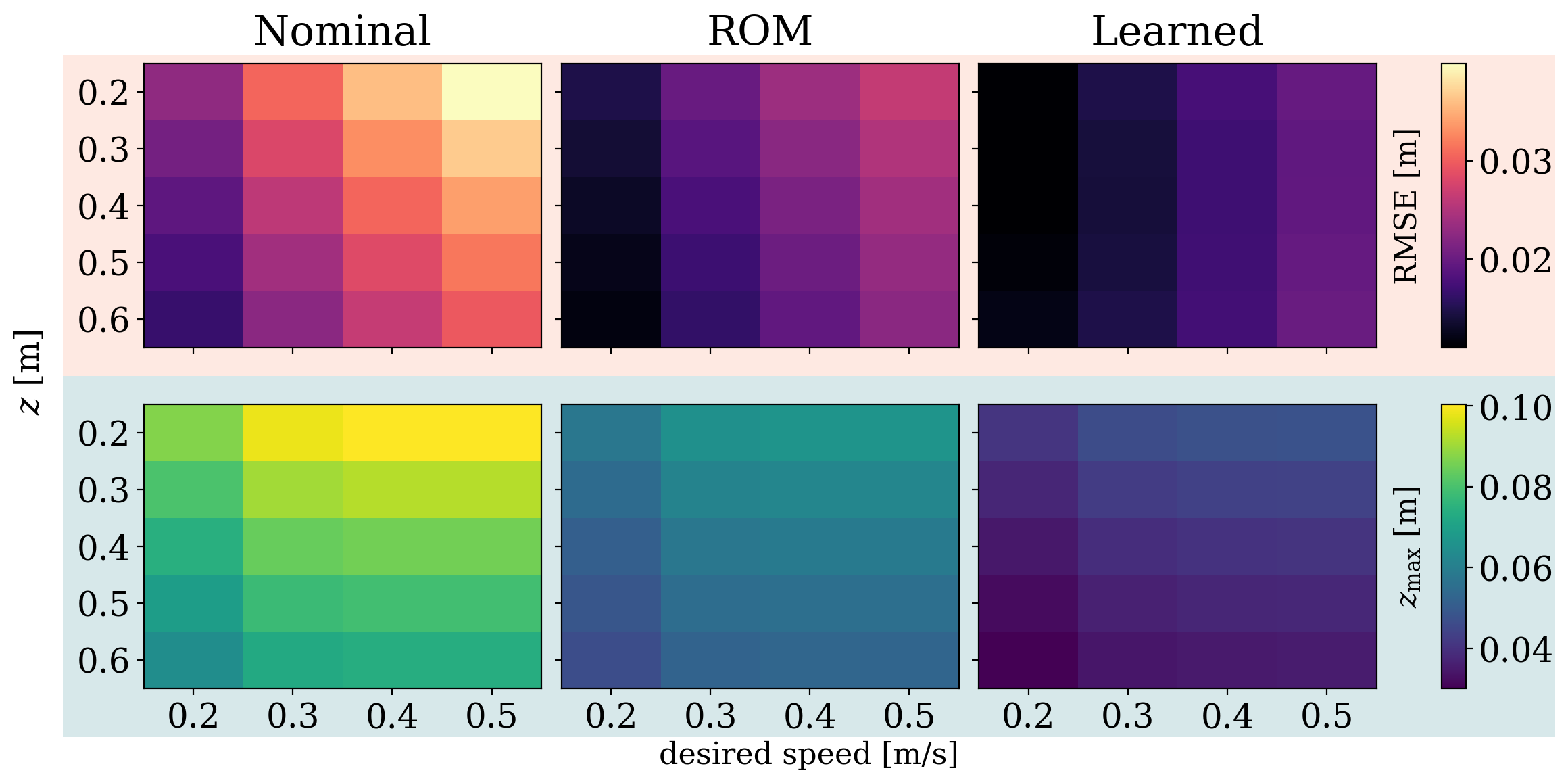}
    \caption{\small Closed-loop tracking performance (RMSE and $z_{max}$) of the FBL controller over varying reference speeds and vertical separations ($z$). The heatmaps illustrate the progressive reduction in tracking error when upgrading from the Nominal model (left) to the ROM (center), and finally to the learned residual (right).}
    \label{fig:systematic-eval-fbl}
    \vspace{-2.5mm}
\end{figure}

\begin{table}[h]
\centering
\setlength{\tabcolsep}{4pt} 
\begin{tabular}{@{}clccc@{}}
\toprule
\textbf{$z$ [m]} & \textbf{Ctrl.} & \textbf{Nominal} & \textbf{ROM} & \textbf{Learned} \\ \midrule
\multirow{2}{*}{0.3} & FBL & 0.030 / 0.090 & 0.020 / 0.061 & \textbf{0.016} / \textbf{0.042} \\
 & SE(3) & 0.088 / 0.157 & 0.054 / 0.096 & \textbf{0.031} / \textbf{0.055} \\ \midrule
\multirow{2}{*}{0.5} & FBL & 0.026 / 0.077 & 0.018 / 0.054 & \textbf{0.016} / \textbf{0.036} \\
 & SE(3) & 0.070 / 0.124 & 0.042 / 0.074 & \textbf{0.022} / \textbf{0.035} \\ \bottomrule
\end{tabular}
\caption{Sim. Tracking Errors (RMSE / $z_{max}$) [m]}
\label{tab:aggregated_control}
\end{table}

Finally, we evaluate our proposed controller against NMPC whose dynamics are augmented with the same learned residual model. Both controllers were tested head-to-head using the previous experimental setups, with results for mean 3D position error and maximum $z$ error reported in Table \ref{tab:nmpc_comparison}. Our flatness-based controller achieves tracking performance on par with the NMPC. Moreover, while the NMPC was implemented in Acados \citep{verschueren2022acados}, compiled into C code, and utilized real-time iteration, our JIT-compiled PyTorch-based controller still achieved a $20\times$ reduction in solve time. This efficiency suggests broader applicability for computationally constrained hardware.

\begin{table}[h]
\centering
\setlength{\tabcolsep}{4pt} 
\begin{tabular}{@{}clcc@{}}
\toprule
\textbf{$z$ [m]} & \textbf{Ctrl.} & \textbf{Error (RMSE/}$z_{max}$\textbf{)} & \textbf{Avg. Solve Time[ms]}  \\ \midrule
\multirow{2}{*}{0.3} & Ours & \textbf{0.016} / 0.042 & \textbf{0.99}  \\
 & NMPC & 0.02 / \textbf{0.032} & 27.6  \\ \midrule
\multirow{2}{*}{0.5} & Ours & \textbf{0.016} / 0.036 & \textbf{1.3}  \\
 & NMPC & 0.018 / \textbf{0.019} & 26.2  \\ \bottomrule
\end{tabular}
\caption{Benchmark against NMPC \citep{chee2024flying}}
\label{tab:nmpc_comparison}
\end{table}
\vspace{-2mm}
\fi

\subsection{Physical Experiments}
\textit{Setup}: 
We use two Crazyflie 2.1 quadrotors equipped with thrust upgrade bundles \citep{bitcraze_crazyflie}. Each quadrotor has a body length of 10 cm and weighs 36 g with the battery and Vicon markers. The Vicon motion capture system streams pose estimates to the base station (a laptop with an Intel i5 CPU) at 120 Hz. This base station computes control commands that are transmitted to the quadrotors at 400 Hz via Crazyradio PA. The software for simultaneous control of the two Crazyflies was implemented using the CrazyROS wrapper \citep{honig2017flying}.


\ifniceformat
\begin{figure}
    \centering
    {\includegraphics[width=0.8\textwidth]{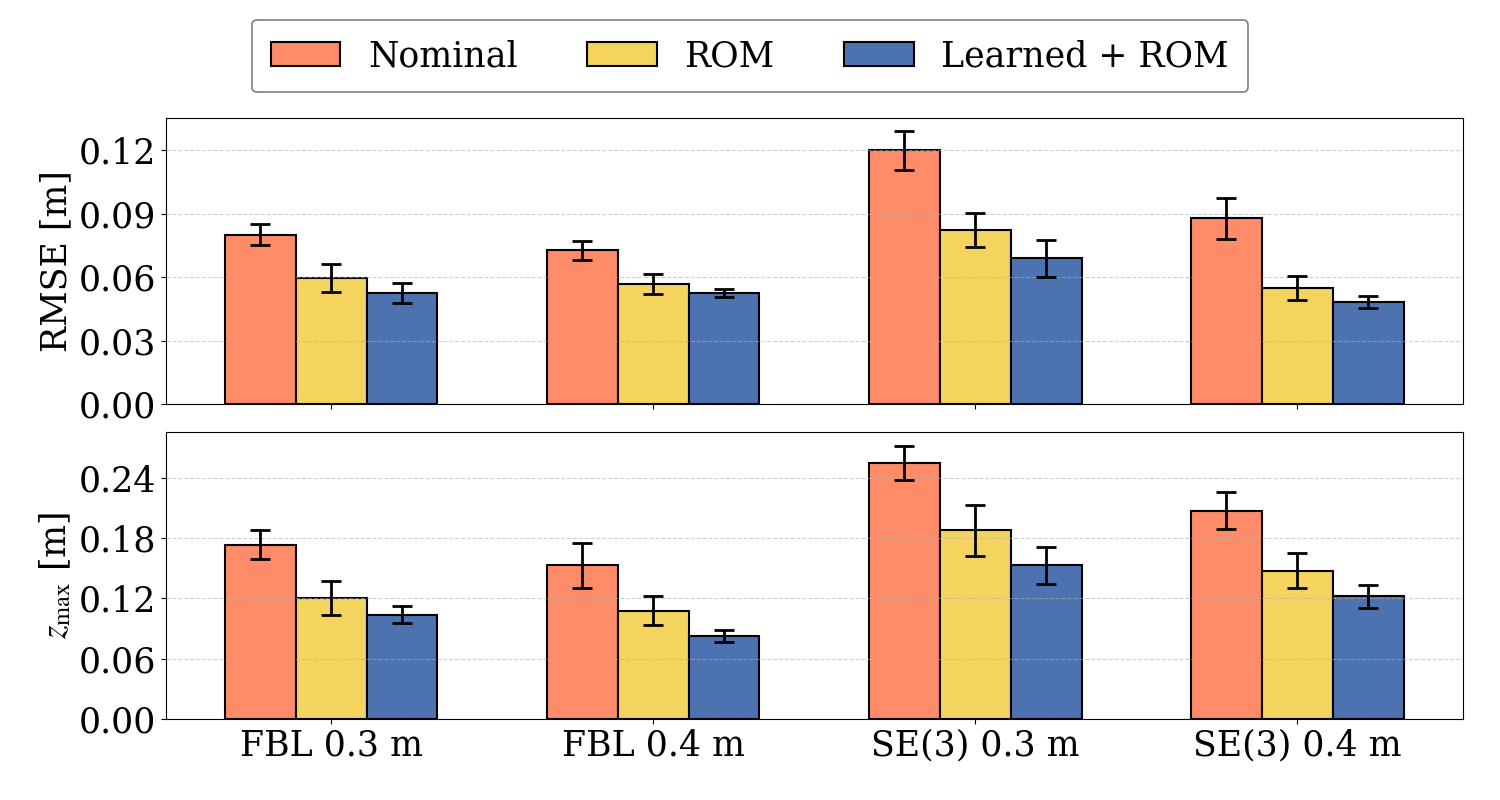}}
    \caption{\small\textbf{Experiment statistics:} Statistics of the proposed tracking controller and the baselines over five runs. The RMSEs are shown in the top subplot, and the bottom subplot illustrates the maximum vertical separation error $(z_{max})$. The mean and standard deviation of the runs are depicted using the bars and error bars. The text and the value in the x-axis labels indicate the controller used and the commanded vertical separation.}
    \label{fig:phy_expt_stats}
\end{figure}
\else
\begin{figure}
    \centering
    {\includegraphics[scale=0.21, trim = 0.9cm 0.5cm 0cm -0.9cm]{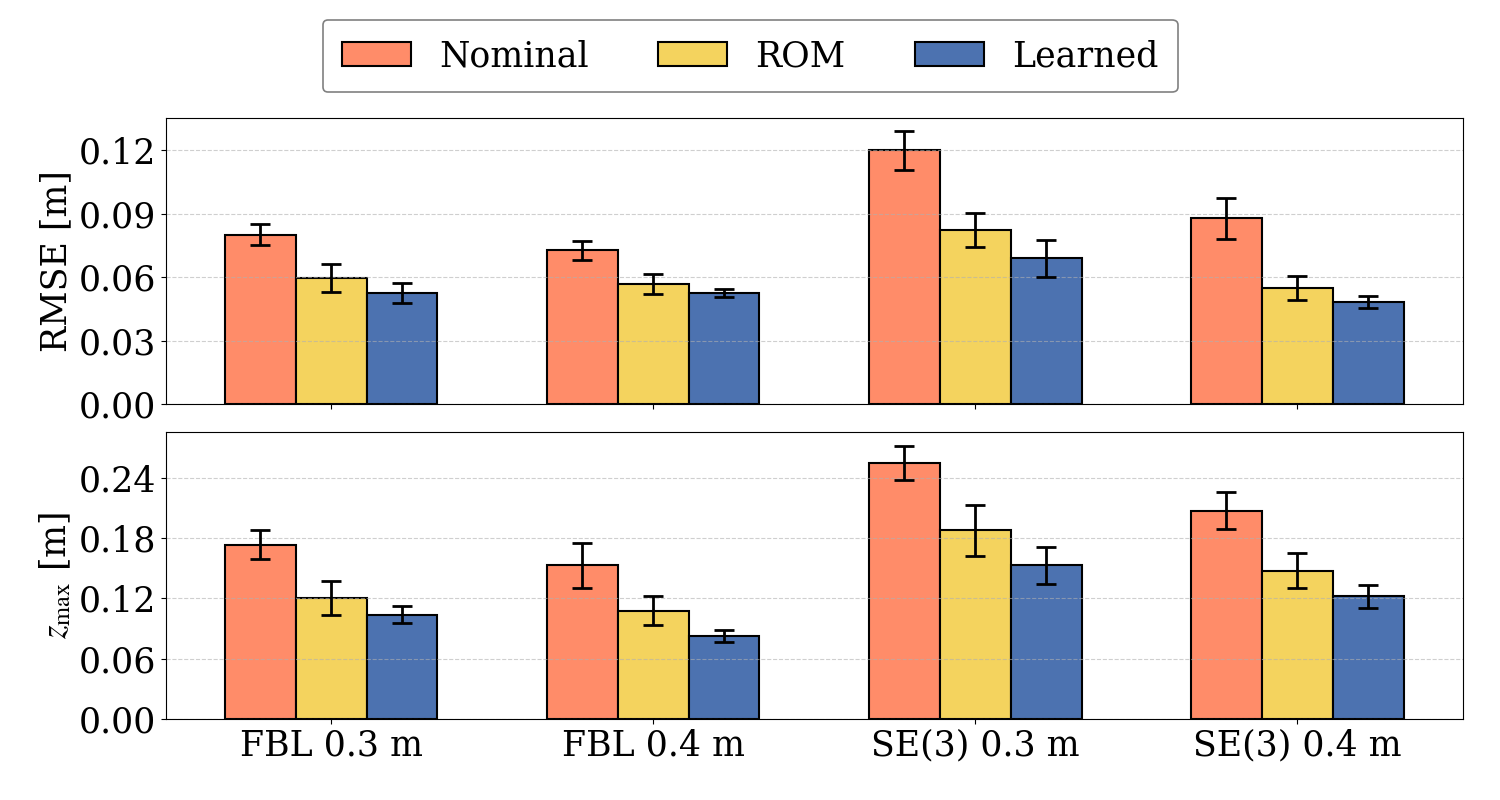}}
    \caption{\small\textbf{Experiment statistics:} Statistics of the proposed tracking controller and the baselines over five runs. The RMSEs are shown in the top subplot, and the bottom subplot illustrates the maximum vertical separation error $(z_{max})$. The mean and standard deviation of the runs are depicted using the bars and error bars. The text and the value in the x-axis labels indicate the controller used and the commanded vertical separation.}
    \label{fig:phy_expt_stats}
    \vspace{-2.5mm}
\end{figure}
\fi

We implemented the FBL and SE(3) controllers using the same software architecture employed in simulation. However, to accommodate the default firmware interfaces, the FBL controller was configured to output thrust and angular velocity commands rather than thrust and torque. Therefore, we compensate exclusively for residual forces and rely on the high-frequency PID controller onboard to handle the residual torques. For the \rewrite{physics-prior}{ROM} downwash model, we set the drag coefficient to $C_D = 0.236$, identified via the system identification procedure in \citet{chee2024flying} to ensure reliable baseline performance. Residual forces are captured using a 2-layer neural network with 4 neurons. The training dataset $\mathcal{D}$ spans 28 seconds (6,720 points) collected from trajectory segments where the quadrotors fly in \textit{static} and \textit{stacked} formations at a 0.4 m separation and 0.4 m/s velocity under an FBL controller unaugmented by learned residual dynamics.


\textit{Results}:
We study the closed-loop performance of both FBL and SE(3) controllers when augmented with residual models of varying fidelities, yielding six controller configurations: Learned-FBL, \rewrite{Prior-FBL}{ROM-FBL}, Nominal-FBL, and their respective SE(3) counterparts. We consider two commanded separations: 0.4m (training distribution) and 0.3m (out-of-distribution). A total of 60 experimental runs were conducted, with each test case repeated five times. As shown in Figure \ref{fig:phy_expt_stats}, our proposed Learned-FBL controller achieved the best overall performance. The \rewrite{Prior-FBL}{ROM-FBL} controller outperformed the Nominal-FBL baseline by an average of $24\%$ in RMSE and $30\%$ in maximum vertical error ($z_{max}$). By leveraging the combined physics-informed and learned framework, the Learned-FBL controller achieved additional improvements of 29\% in RMSE and 43\% in $z_{max}$ relative to \rewrite{Prior-FBL}{ROM-FBL}. A similar trend was observed when integrating the residual models as feedforward terms into the SE(3) controllers: the Learned-SE(3) and \rewrite{Prior-SE(3)}{ROM-SE(3)} configurations reduced RMSE by 43.7\% and 34\%, respectively, compared to the nominal baseline. Notably, these performance gains were realized without any additional data collection or retraining of the neural network for the SE(3) variants. The superior performance of controllers utilizing high-fidelity models over their lower-fidelity counterparts indicates that our physics-informed parameterization successfully captures the complex aerodynamic interactions inherent in tight formation flight.

Overall, our Learned-FBL controller achieved an average RMSE of 5 cm and a $z_{max}$ of 10 cm. These results match the reported tracking performance of KNODE-DW MPC \citep{chee2024flying}, a state-of-the-art NMPC framework that also utilizes learned residual models. Crucially, our flatness-based approach achieves this state-of-the-art accuracy while avoiding the substantial computational overhead of real-time NMPC optimization. Finally, to demonstrate the agility unlocked by our computationally efficient method, we challenge our proposed controller to merge from a 1.1 m stacked formation into a tight 0.1 m vertical separation to fly through a 0.4 m high window (Figure \ref{fig:cover_image}).

\section{Conclusion}\label{sec:conclusion}
We introduced a computationally efficient framework for parameterizing and learning residual dynamics by leveraging differential flatness and physics-based \rewrite{priors}{ROM}. We presented a flatness-based controller that utilizes these learned residuals to achieve tracking performance on par with NMPC baselines, while requiring an order of magnitude less computation. Through both simulation and hardware experiments, we demonstrated that our method effectively handles complex multi-quadrotor proximity flight scenarios with significant performance gains over baseline controllers. Future work includes relaxing the structural assumptions of the flatness-preserving residuals and scaling up the experiments.


\ifniceformat
\bibliographystyle{unsrtnat}
\else
\bibliographystyle{bibFiles/IEEEbib}
\fi
\bibliography{bibFiles/sample}

\appendix
\ifniceformat
\section{Details on the Physics Prior}\label{sec:physics-prior-details}
\else
\subsection{Details on the \rewrite{Physics Prior}{Reduced-order Model}}\label{sec:physics-prior-details}
\fi
Approximating the downwash with turbulent jet theory \citep{bauersfeld2024robotics, kiran2024downwash} yields induced airflow velocity:
\begin{equation}
\label{eq:aero_dis}
V\left(\Delta z, \Delta r\right) = \frac{U_H \frac{Bd}{\Delta\tilde{z} - z_0}
}{\left( 1 + \left(\sqrt{2} - 1\right) \left(\frac{\Delta \tilde{r}}{S\left(\Delta \tilde{z} - z_0\right)}
\right)^2 \right)^2}\vce_z, 
\end{equation}
where $\Delta \tilde{z} := \Delta z/ \lambda$ and $\Delta \tilde{r} := \Delta r/ \lambda$ denote the vertical and horizontal separation normalized over the half-body length $\lambda$ of the quadrotor. The constants $B_d$, $S$, $U_H$, and $z_0$ characterize the turbulent jet and are obtained through empirical airflow velocity measurements \citep{bauersfeld2024robotics}. By modeling the \textit{bottom} quadrotor as a disk, the downwash-induced forces and torques are derived by integrating the drag equations over the disk's surface area $A_b$ as follows \citep{jain2019modeling}: 
\begin{subequations}
\begin{align}
\hat f^{a}_{d}\left(\vcp^1, \vcp^2\right) &= R^{\wdf}_{\mathcal{F}}(\int_{A_b} \, C_D \, \zeta \, dA)\vce_z, \\
\hat f^{\tau}_{d}\left(\vcp^1, \vcp^2\right) &= R^{\wdf}_{\mathcal{F}}\int_{A_b} \Big( \vcp_{A_b} - \vcp' \Big) \times  \vce_z\, C_D \, \zeta \, dA,
\end{align}
\end{subequations}
where $\vcp' =R^{\mathcal{F}}_{\wdf} \Delta \vcp$, and $\vcp_{A_b}\in\mathbb{R}^3$ denotes the position of the points on $A_b$ in $\mathcal{F}$. The constant $C_D \in \mathbb{R}$ denotes the drag coefficient, and $\zeta := \frac{1}{2} \rho ||V(\Delta z, \Delta r)||_2^2 \in \mathbb{R}$ represents the dynamic pressure.
\end{document}